\documentclass[runningheads]{llncs}
\usepackage{caption}
\usepackage{graphicx}
\usepackage{url}
\usepackage{xcolor}
\usepackage{pifont}
\usepackage{amsmath}
\usepackage{amssymb}
\usepackage{multirow}
\usepackage[ruled,vlined,linesnumbered]{algorithm2e}

\definecolor{myRed}{rgb}{0.808,0.067,0.149}
\definecolor{myGreen}{rgb}{0.067,0.708,0.149}

\usepackage{xcolor}

\newcommand{\cmark}{{\color{myGreen}\ding{51}}}

\begin{document}
\title{Learning Using Generated Privileged Information by Text-to-Image Diffusion Models}
\titlerunning{Learning Using Generated Privileged Information}
%
\author{Anonymous Author(s)}
\institute{Anonymous Institute}

\author{Rafael-Edy Menadil \and Mariana-Iuliana Georgescu \and Radu Tudor Ionescu\thanks{Corresponding author: \email{raducu.ionescu@gmail.com}}}
\institute{Department of Computer Science, University of Bucharest, Bucharest, Romania}
\authorrunning{R.E. Menadil et al.}
%
\maketitle              
\begin{abstract}
Learning Using Privileged Information is a particular type of knowledge distillation where the teacher model benefits from an additional data representation during training, called privileged information, improving the student model, which does not see the extra representation. However, privileged information is rarely available in practice. To this end, we propose a text classification framework that harnesses text-to-image diffusion models to generate artificial privileged information. The generated images and the original text samples are further used to train multimodal teacher models based on state-of-the-art transformer-based architectures. Finally, the knowledge from multimodal teachers is distilled into a text-based (unimodal) student. Hence, by employing a generative model to produce synthetic data as privileged information, we guide the training of the student model. Our framework, called Learning Using Generated Privileged Information (LUGPI), yields noticeable performance gains on four text classification data sets, demonstrating its potential in text classification without any additional cost during inference. 
\end{abstract}

\setlength{\abovedisplayskip}{3.5pt}
\setlength{\belowdisplayskip}{3.5pt}

\vspace{-0.1cm}
\section{Introduction}
\vspace{-0.2cm}

In the quest of developing effective and efficient machine learning models, researchers developed the knowledge distillation framework \cite{Ba-NIPS-2014,Hinton-DLRL-2014}, in which the outputs of one \cite{Ba-NIPS-2014,Park-CVPR-2019} or more \cite{Hinton-DLRL-2014,You-KDD-2017,Yu-CVPR-2019} typically heavy models, called \emph{teachers}, are used as target for a typically lightweight model, called \emph{student}. This framework is primarily used to compress very deep models into shallower, yet effective models \cite{Feng-arxiv-2019,Lopez-ICLR-2016,Park-CVPR-2019,Yim-CVPR-2017,You-KDD-2017,Yu-CVPR-2019}. A secondary use of knowledge distillation is to leverage additional data representations, available only at training time, to improve the performance of a model which does not have access to the extra representation. This latter framework, called Learning Using Privileged Information (LUPI) \cite{Vapnik-NN-2009}, was introduced well before the era of deep learning, but it was later shown \cite{Lopez-ICLR-2016} that it represents a particular kind of knowledge distillation.

\begin{figure}[!th] 
\centering
\includegraphics[width=\textwidth]{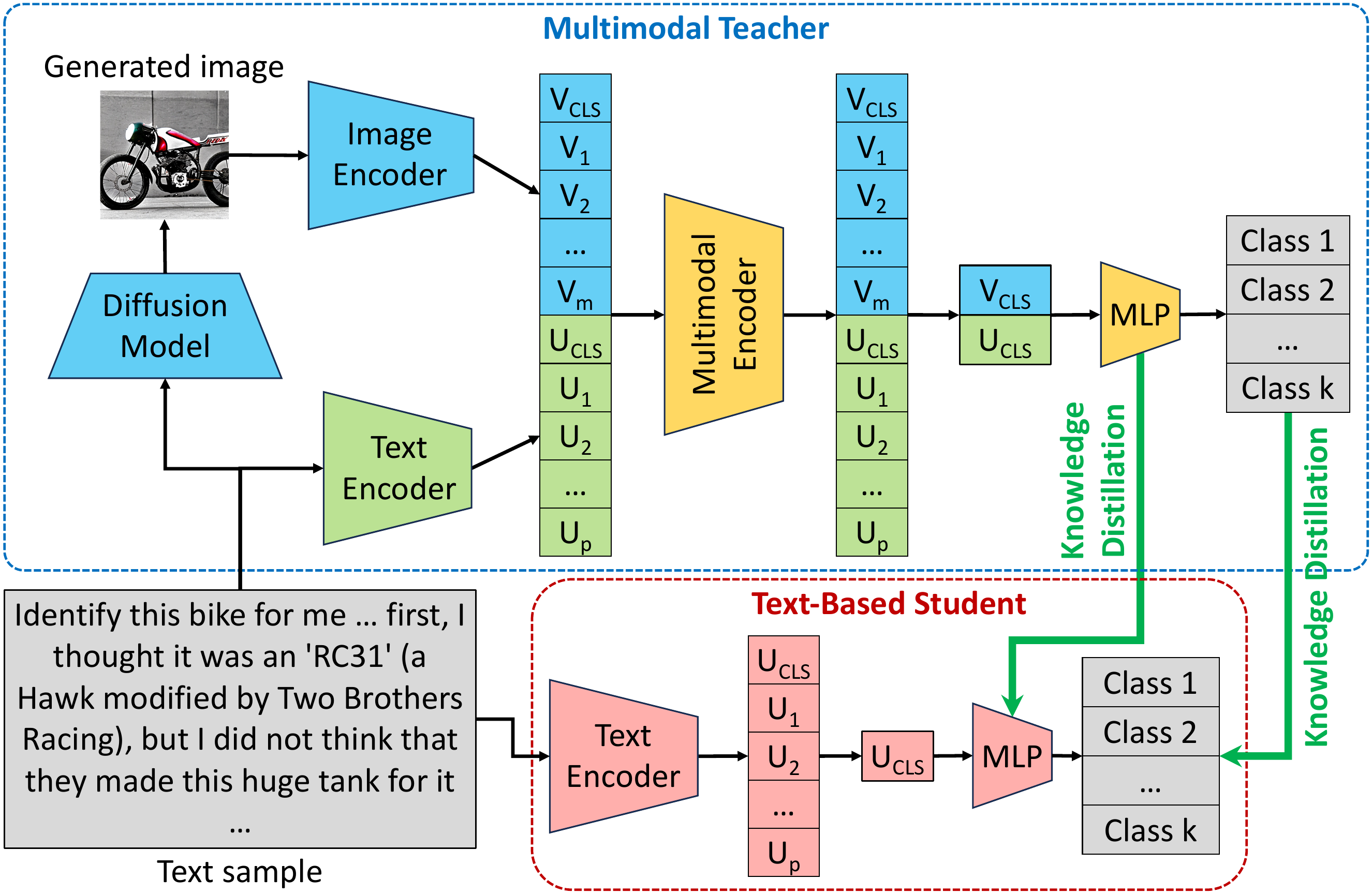}
\vspace{-0.5cm}
\caption{An illustration of our Learning Using Generated Privileged Information (LUGPI) framework. For each text sample, a diffusion model generates an image. The original text sample and the generated image are used to train a multimodal teacher model. Then, a text-based student model is trained via knowledge distillation from the teacher. The distillation is carried out at two levels.}\label{fig:pipeline}
\vspace{-0.3cm}
\end{figure}

Although LUPI is an interesting and useful framework, it has rarely been applied in solving mainstream machine learning problems \cite{Alehdaghi-ECCVW-2022,Georgescu-MVA-2022,Georgescu-ICPR-2021,Jung-NeurIPS-2022,Vapnik-NN-2009,Zhao-PR-2022}, since finding additional modalities to represent the training data is not an easy task. With the advent of diffusion models \cite{Croitoru-TPAMI-2023,Ho-NeurIPS-2020,Sohl-ICML-2015,Song-NeurIPS-2019}, which demonstrated impressive capabilities in generating realistic and diverse images based on text prompts \cite{Avrahami-CVPR-2022,Gu-CVPR-2022,Rombach-CVPR-2022,Saharia-NeurIPS-2022}, we can now automatically generate image representations of text samples without much effort. To this end, we propose a novel framework called Learning Using Generated Privileged Information (LUGPI), which harnesses a state-of-the-art text-to-image diffusion model to generate the privileged information, namely Stable Diffusion v2 \cite{Rombach-CVPR-2022}. Our framework is applied on text classification tasks, where the original modality is represented by text samples and the additional modality is represented by images. Next, we train multimodal teacher models based on combining state-of-the-art transformer-based architectures, such as Distilled Bidirectional Encoder Representations from Transformers
(DistilBERT) \cite{Sanh-EMC-2019}, Vision Transformer (ViT) \cite{Dosovitskiy-ICLR-2021} and Contrastive Language-Image Pre-Training (CLIP) \cite{Radford-ICML-2021}. Remarkably, we find that our multimodal teachers outperform the standalone text-based (unimodal) model. However, employing the multimodal teachers during inference would inherently imply the use of the diffusion model to generate the images. This greatly impacts the inference time of the whole framework, since diffusion models are notoriously known for being computationally expensive \cite{Croitoru-TPAMI-2023}. For instance, Stable Diffusion v2 \cite{Rombach-CVPR-2022} comprises about 865 million learnable parameters, requiring about 17 seconds to generate a single image on an NVIDIA GeForce RTX 3090 24GB GPU. To address this limitation, we distill the knowledge from a multimodal teacher into a text-based student model, as shown in Figure \ref{fig:pipeline}. This completely eliminates the need to generate images during inference. Thus, LUGPI does not increase the computational cost at test time. 

We carry out experiments on four text classification data sets to evaluate the proposed framework and compare it with the conventional training approach  based on pre-training and fine-tuning, while preserving the underlying DistilBERT architecture \cite{Sanh-EMC-2019}. Our empirical results indicate that LUGPI brings significant performance gains on all four data sets.

In summary, our contribution is threefold:
\begin{itemize}
    \item \vspace{-0.2cm} We propose to harness diffusion models in order to artificially generate an extra data modality in the form of images, complementing the text modality, which enables us to train more powerful multimodal neural models.
    \item We introduce the novel Learning Using Generated Privileged Information framework to distill knowledge from our multimodal teachers into text-based (unimodal) students.
    \item We conduct experiments on four benchmarks, showing that the proposed framework improves the accuracy rates of text-based models by noticeable margins, without any extra cost during inference. 
\end{itemize}

\vspace{-0.2cm}
\section{Related Work}
\vspace{-0.2cm}

\noindent
\textbf{Learning Using Privileged Information.}
There are two types of knowledge distillation frameworks, which were independently introduced in literature, namely model compression \cite{Ba-NIPS-2014,Hinton-DLRL-2014} and learning using privileged information \cite{Vapnik-NN-2009}. In 2016, Lopez-Paz et al.~\cite{Lopez-ICLR-2016} unified the model compression and learning under privilege information paradigms into the knowledge distillation framework.

The model compression technique \cite{Ba-NIPS-2014,Hinton-DLRL-2014}  is mainly aimed at training a shallow and efficient student architecture using one or more deeper and powerful teachers. In this way, a shallow student could benefit from the knowledge gained by a deep teacher, while having less parameters, and consequently, a lower running time during inference. 

The learning using privileged information paradigm \cite{Vapnik-NN-2009} was introduced to transfer the knowledge from a teacher model, which is trained with privileged information, to a student model, which does not have access to the privileged data. In this scenario, the teacher and the student can share the same architecture, the main difference being the data used to train the two models. Many recent works \cite{Alehdaghi-ECCVW-2022,Gao-MIA-2019,Garcia-TPAMI-2019,Georgescu-ICPR-2021,Georgescu-MVA-2022,Lee-ECCV-2020,Liu-CBM-2023,Yuan-Arxiv-2018} applied the LUPI framework to improve the performance of the student without using additional information at test time. For example, Yuan et al.~\cite{Yuan-Arxiv-2018} trained a student to estimate the 3D hand pose using only the RGB image at test time. The knowledge about the depth channel was transferred from the teacher during the knowledge distillation process. Alehdaghi et al.~\cite{Alehdaghi-ECCVW-2022} decreased the gap between RGB and infrared images used in the person re-identification task by applying the LUPI framework. They proposed to create an intermediate virtual domain that acts as a bridge between the two image modalities.  The intermediate virtual domain was used as privileged information for the student model during training. Georgescu et al.~\cite{Georgescu-ICPR-2021} applied LUPI for facial expression recognition under strong occlusion, where the teacher learns from completely visible faces, but the student can only use occluded faces as input. They later extended their approach to age estimation and gender prediction from faces \cite{Georgescu-MVA-2022}.

Similar to the aforementioned works~\cite{Alehdaghi-ECCVW-2022,Gao-MIA-2019,Garcia-TPAMI-2019,Georgescu-ICPR-2021,Georgescu-MVA-2022,Lee-ECCV-2020,Liu-CBM-2023,Yuan-Arxiv-2018}, we use extra data as privileged information during training. Different from the related studies on LUPI, our method does not require the existence of additional representations, since it generates the privileged data using a generative diffusion model. Hence, our framework broadens the applicability of LUPI to text-based corpora that do not have additional representations of the data samples.

\noindent
\textbf{Data augmentation.} 
Our approach can also be seen as a rather unconventional data augmentation technique. However, data augmentation is usually employed to improve the robustness to data variation~\cite{Devries-Arxiv-2017,Zhang-ICLR-2018}, while in our case, we employ it to obtain privileged information. In general, data augmentation plays an important role in increasing the performance of deep learning architectures~\cite{Devries-Arxiv-2017,Zhang-ICLR-2018}, especially when the available training data is limited. The most common data augmentation methods used in computer vision are methods based on rotating, cropping and flipping the images~\cite{Cubuk-NEURIPS-2020}. Although techniques like these can offer better performance than just training on the original data, they lack the capability of creating a completely different data point, instead relying on the existing data and manipulating it just enough to have a variety within the training data. 

In recent years, we have seen generative models, such as Generative Adversarial Networks (GANs) \cite{Goodfellow-NIPS-2014} and diffusion models, that have been used to successfully augment data and improve the accuracy of various models \cite{Antoniou-ICANN-2018,Azizi-arxiv-2023,Qian-SC-2019,Sandfort-Nature-2019}. Generative models can create new data points that closely resemble the training data distribution, often being mistaken with natural data points. Therefore, classification models can leverage this new data variety to offer high performance without having to gather any new data points. 
Furthermore, there are some examples that successfully use generative models when conventional techniques fall short \cite{Shivashankar-CVPRW-2023,Yang-arxiv-2023}. Yang et al.~\cite{Yang-arxiv-2023} proposed to use diffusion models to generate images illustrating human-object interactions, conditioned by prompts explaining the interactions. Shivashankar et al.~\cite{Shivashankar-CVPRW-2023} trained a GAN model to generate images along with their segmentation label for medical and face segmentation data sets. In these cases, conventional data augmentation methods provide suboptimal results when compared with generative models. This is because the latter models can generate new data points that resemble the training data distribution, aside from being able to generate variations of existing data points conditioned by some specific features that need to be present in the generated output.

Unlike other data augmentation techniques, we propose to generate image-based representations from text samples, essentially obtaining a new modality. Thus, our technique requires employing multimodal models to benefit from the extra data representation. To return to using a unimodal input while keeping the benefits of the multimodal data, we employ knowledge distillation.

\vspace{-0.2cm}
\section{Method}
\vspace{-0.2cm}

\noindent
\textbf{Overview and motivation.}
Learning Using Privileged Information \cite{Vapnik-NN-2009} is suitable for machine learning tasks where the training data is represented by multiple modalities. However, the majority of machine learning problems only involve a single modality, rendering LUPI inapplicable. To overcome this challenge in the area of natural language processing and text classification, we propose to utilize a text-to-image diffusion model to generate privileged information in the form of images, in order to solve text classification problems where privileged information is not typically available. 

We believe that our proposal is grounded in how the human mind works. For instance, humans use their imagination to mentally visualize objects, colors, textures or other visual aspects evoked in a text. This process helps humans in reaching a better and deeper text comprehension \cite{Gambrell-RRQ-1993}. In a similar way, we conjecture that imaginary pictures can boost the performance of neural models such as BERT \cite{Devlin-NAACL-2019} or DistilBERT \cite{Sanh-EMC-2019}, provided that the visualizations are sufficiently representative. To increase the chances of successfully implementing our proposal, we make use of diffusion models, which are considered by many researchers as state-of-the-art text-to-image generators \cite{Dhariwal-NeurIPS-2021}, surpassing previous models based on GANs.

To harness the generated images, a straightforward approach is to employ models on both text and image modalities in order to improve text classification performance. However, this approach is suboptimal in terms of speed, requiring additional time to generate and process images during inference. Our framework addresses this issue through knowledge distillation, i.e.~the knowledge learned by the multimodal model, called teacher, is distilled into a text-based model, called student. At test time, we employ the student model to make predictions, thus eliminating the need to generate and process images. Our training framework is formally introduced in Algorithm \ref{alg:LUGPI}. We first introduce the notations, then continue by presenting the three stages of our algorithm, namely image generation, teacher model training and knowledge distillation. 

\begin{algorithm*}[!t]
\caption{Learning Using Generated Privileged Information}
\label{alg:LUGPI}
\small{
\KwIn{$\mathcal{D}$ - the training set of labeled text samples, $G$ - the text-conditional diffusion model, $\theta_G$ - the weights of the diffusion model, $T$ - the multimodal teacher model, $S$ - the student model, $\theta^*_T$ - (optional) pre-trained weights for the teacher, $\theta^*_S$ - (optional) pre-trained weights for the student, $\eta_T$ - the teacher's learning rate, $\eta_S$ - the student's learning rate, $\alpha$ - the importance of the cross-entropy between the teacher and the student, $\beta$ - the importance of the mean squared error between the teacher and student embeddings.}
\KwOut{
$\theta_S$ - the trained weights of the student model.
}
$n \leftarrow \lvert \mathcal{D} \rvert;$ $\lhd$ get the number of training samples\\ 
$X' \leftarrow \emptyset;$ $\lhd$ initialize the set of generated images\\
\ForEach{$i \in \{1,2,...,n\}$}
{
    $x'_i \leftarrow G(x_i, \theta_G);$ $\lhd$ generate an image for the text sample $x_i$\\
    $X' \leftarrow X' \cup \{ x'_i \};$ $\lhd$ add the generated image to the set $X'$\\
}
\If{$\theta^*_T \neq \emptyset$}{
$\theta_T \leftarrow \theta^*_T;$ $\lhd$ initialize weights of teacher using pre-trained weights\\
}
\Else{
$\theta_T \sim \mathcal{N}\left(0,\frac{2}{d_{in} + d_{out}}\right);$ $\lhd$ initialize weights of teacher using Xavier init \cite{Glorot-AISTATS-2010}\\
}
\Repeat{convergence}{
        \ForEach{$i \in \{1,2,...,n\}$}{
            $t_i \leftarrow T(x_i, x'_i, \theta_T);$ $\lhd$ get class probabilities predicted by the teacher\\
            $\theta_T \leftarrow \theta_T - \eta_T \cdot \nabla \mathcal{L}_{\scriptsize{\mbox{CE}}}(y_i, t_i);$ $\lhd$ train the teacher using cross-entropy\\
        }
    }
    
\If{$\theta^*_S \neq \emptyset$}{
$\theta_S \leftarrow \theta^*_S;$ $\lhd$ initialize weights of student using pre-trained weights
}
\Else{
$\theta_S \sim \mathcal{N}\left(0,\frac{2}{d_{in} + d_{out}}\right);$ $\lhd$ initialize weights of student using Xavier init \cite{Glorot-AISTATS-2010}\\
}
\Repeat{convergence}{
        \ForEach{$i \in \{1,2,...,n\}$}{
            $t_i, e^T_i \leftarrow T(x_i, x'_i, \theta_T);$ $\lhd$ get probabilities and embedding from teacher\\
            $s_i, e^S_i \leftarrow S(x_i, \theta_S);$ $\lhd$ get probabilities and embedding from student\\
            $\mathcal{L}_{\scriptsize{\mbox{KD}}} \leftarrow \mathcal{L}_{\scriptsize{\mbox{CE}}}(y_i, s_i) + \alpha \cdot \mathcal{L}^T_{\scriptsize{\mbox{CE}}}(t_i, s_i) + \beta \cdot \mathcal{L}^T_{l_2}(e^T_i, e^S_i);$ $\lhd$ apply Eq.~\eqref{eq_loss_KD}\\
            $\theta_S \leftarrow \theta_S - \eta_S \cdot \nabla \mathcal{L}_{\scriptsize{\mbox{KD}}};$ $\lhd$ train the student using the joint loss\\
        }
    }
}
\end{algorithm*}

\noindent
\textbf{Notations.}
Let $\mathcal{D} = (X,Y) = \{(x_1, y_1), (x_2, y_2), ..., (x_n, y_n)\}$ represent a training set of text samples, where $n$ is the number of samples in the data set, and $y_i$ is the ground-truth label associated with text sample $x_i$. Let $T$ and $\theta_T$ represent the multimodal teacher model and its weights, respectively. Similarly, let $S$ and $\theta_S$ represent the text-based student model and its weights. The weights of the teacher and student models are updated using the learning rates $\eta_T$ and $\eta_S$, respectively. Let $X' = \{x'_1, x'_2, ...,x'_n \}$ represent the set of images generated by a diffusion model $G$ with the weights $\theta_G$. Let $\mathcal{N}(\mu,\sigma^2)$ represent the normal distribution of mean $\mu$ and standard deviation $\sigma$. Let $e^T_i$ and $e^S_i$ denote the embedding vectors produced by the teacher and the student for the $i$-th data sample, respectively. The embedding vectors are taken just before the classification layer of each model.

\noindent
\textbf{Image generation.} In steps 2-5 of Algorithm \ref{alg:LUGPI}, we utilize a pre-trained text-to-image diffusion model to generate privileged information in the form of images. In step 4, the generator $G$ generates an image denoted by $x_i'$ conditioned on the text sample $x_i$. In step 5, the generated image is added to the set $X'$. Steps 4 and 5 are repeated until all training examples are passed through $G$.

We choose the Stable Diffusion v2 \cite{Rombach-CVPR-2022} model trained on the LAION-5B \cite{Schuhmann-NeurIPS-2022} data set as our generator $G$. The use of this model is chosen in favor of another open-source diffusion model, namely GLIDE \cite{Nichol-ICML-2021}. To decide on which generator to use, we visually inspected their outputs on a subset of 100 prompts from the chosen data sets. We observed that Stable Diffusion v2 is usually better aligned with the provided text prompts than GLIDE. This influenced our decision towards using the former model.


\noindent
\textbf{Teacher training.} The second stage of our pipeline is dedicated to training the teacher model. This stage corresponds to steps 6-14 of Algorithm \ref{alg:LUGPI}. The teacher model is a multimodal architecture comprising three transformer-based encoders: a text encoder, an image encoder, and a multimodal encoder. As illustrated in Figure \ref{fig:pipeline}, the tokens produced by the text encoder are concatenated with the tokens given by the image encoder. The concatenated set of tokens is further passed through the multimodal encoder, which comprises a vanilla transformer block based on multi-head attention, having 8 attention heads. The multimodal encoder learns to perform cross-modal attention, strengthening relations across the text and image modalities. From the resulting set of multimodal tokens, we keep the classification token $U_{\scriptsize{\mbox{CLS}}}$ from the text modality and the classification token $V_{\scriptsize{\mbox{CLS}}}$ from the image modality, discarding the other tokens. This is a conventional procedure when transformers are applied to downstream classification tasks \cite{Devlin-NAACL-2019,Dosovitskiy-ICLR-2021}. Next, the classification tokens are concatenated and given as input to a multi-layer perceptron (MLP) with two layers, where the first layer comprises 786 neurons and the second one comprises $k$ neurons, where $k$ is the number of classes. A softmax function computes the output probabilities.

In order to make the prediction $t_i$, the teacher model $T$ takes the text sample $x_i$ and the generated image $x'_i$ as input, according to step 12 of Algorithm \ref{alg:LUGPI}. In step 13, the weights of the teacher $\theta_T$ are updated using gradient descent, where the gradient is computed with respect to the cross-entropy loss. For the vector of predicted class probabilities $t_i$ and the one-hot label encoding $y_i$, the cross-entropy loss is given by:
\begin{equation}\label{eq_loss_ce}
   \mathcal{L_{\scriptsize{\mbox{CE}}}}(y_i, t_i) = - \sum_{j=1}^k y_{ij} \cdot \log(t_{ij}), \forall i \in \{1,2,...,n \},
\end{equation}
where $k$ is the number of classes.

In our implementation, we choose to use pre-trained architectures for the text and image encoders. For a fair and representative evaluation, we use the same text encoder as the baseline and the student models, namely DistilBERT \cite{Sanh-EMC-2019}. This is to ensure that the observed performance gains are not due to the use of a more powerful text encoder for the teacher model, but rather due to the extra image modality. For the image encoder, we consider two alternative architectures, namely ViT \cite{Dosovitskiy-ICLR-2021} and CLIP Image \cite{Radford-ICML-2021}.

\noindent
\textbf{Knowledge distillation.}
After training the teacher, we apply the knowledge distillation procedure to transfer the knowledge from the multimodal teacher to the student. This stage corresponds to steps 15-25 of Algorithm \ref{alg:LUGPI}. According to steps 15-18, the student can optionally be pre-trained in a standard fashion, prior to the knowledge distillation procedure. We utilize this option to ensure a fair comparison with the baseline model. More precisely, both the baseline DistilBERT and our student DistilBERT are pre-trained. In general, when there are no pre-trained weights for the student, we can simply initialize the model using a conventional approach (step 18), such as Xavier initialization \cite{Glorot-AISTATS-2010}.

The student model is jointly optimizing three objectives. On the one hand, the student has to minimize the cross-entropy loss with respect to the ground-truth (hard) labels, to ensure that its predictions are correct. On the other hand, the student has to optimize the cross-entropy with respect to the probabilities (soft labels) predicted by the teacher, as well as minimize the mean squared error between the corresponding embeddings produced by the teacher and the student, which enables the student to learn knowledge from the teacher model. Formally, for the $i$-th data sample, the joint objective is computed as follows:
\begin{equation}\label{eq_loss_KD}
\begin{split}
\mathcal{L}_{KD} &= \mathcal{L}_{\scriptsize{\mbox{CE}}}(y_i, s_i) + \alpha \cdot \mathcal{L}^T_{\scriptsize{\mbox{CE}}}(t_i, s_i) + \beta \cdot \mathcal{L}^T_{l_2}(e^T_i, e^S_i)\\
&=\! -\! \sum_{j=1}^k y_{ij}\!\cdot\!\log(s_{ij})\! - \!\alpha\!\cdot\! \sum_{j=1}^k t_{ij}\!\cdot\!\log(s_{ij})\!+\! \beta\!\cdot\!\lVert e^T_i - e^S_i \rVert^2_2, \forall i \in \{1,...,n\},\\
\end{split}
\end{equation}
where $\alpha, \beta \geq 0$ are two hyperparameters that control the importance of the knowledge distillation objectives. Note that the distillation is carried out at two levels, namely with respect to the embedding space and the output space. Our ablation study shows the importance of distilling knowledge at both levels.

\vspace{-0.2cm}
\section{Experiments}
\vspace{-0.2cm}

We conduct experiments on four data sets covering three tasks: opinion mining, text categorization by topic, and complex word identification. The data sets are chosen to provide a comprehensive evaluation of image generation and privileged information in different target tasks.

\subsection{Data Sets}
\vspace{-0.2cm}

\noindent
\textbf{IMDB Large Movie Review.}
The IMDB Large Movie Review data set \cite{Maas-ACL-2011} is a well-known benchmark for polarity classification, which is composed of 50,000 movie reviews separated into 25,000 for training and 25,000 for testing. We keep $10\%$ of the training set for validation purposes. The scope of this data set is to predict the polarity of the sentiment (positive or negative).

\noindent
\textbf{20 Newsgroups.}
The 20 Newsgroups data set \cite{Lang-ICML-1995} is a popular benchmark for text categorization by topic. It comprises 18,828 documents that are assigned to one of 20 different categories, ranging from technology to sports and religion. 
In our experiments, we divide the data set into 11,353 training documents, 1,261 validation documents and 6,214 test documents.

\noindent
\textbf{English News.} The English News corpus \cite{Yimam-RANLP-2017} comprises 17,861 sentences with marked words or multi-word phrases that are annotated with complexity levels by native and non-native English speakers. The task is to determine if the target words or multi-word phrases are complex or not. The corpus is divided into 14,002 training sentences, 1,764 validation sentences and 2,095 test sentences.

\noindent
\textbf{English WikiNews.} Another corpus for complex word identification introduced by Yimam et al.~\cite{Yimam-RANLP-2017} is English WikiNews. It has a similar format to English News. The English WikiNews data set is divided into 7,746 training sentences, 870 validation sentences and 1,287 test sentences.

\vspace{-0.2cm}
\subsection{Experimental Setup}
\vspace{-0.2cm}

\begin{table}[!t]
\caption{Accuracy rates on IMDB Large Movie Review \cite{Maas-ACL-2011}, 20 Newsgroups \cite{Lang-ICML-1995}, English News \cite{Yimam-RANLP-2017} and English WikiNews \cite{Yimam-RANLP-2017} data sets. Our teacher and student models are compared with the fine-tuned vanilla DistilBERT \cite{Sanh-EMC-2019}. For reference, we report results with the independent image encoders, namely ViT \cite{Dosovitskiy-ICLR-2021} and CLIP \cite{Radford-ICML-2021}. The best accuracy on each corpus is highlighted in bold. Significantly better results (at a p-value of 0.001) based on McNemar / Cochran Q testing are marked with $\ddagger$.}
\label{tab1}
\centering
\begin{tabular}{|l|c|c|c|c|c|c|}
\hline
\multirow{2}{*}{Model} & \multicolumn{2}{c|}{Modality} & {IMDB} & {20 News} & English & English\\
\cline{2-3}
     & {Text} & {Image} & Reviews & groups & News & WikiNews\\
\hline
\hline
DistilBERT \cite{Sanh-EMC-2019}                     
& \cmark 
& & 0.919 & 0.918 & 0.861 &  0.842 \\
\hline
ViT \cite{Dosovitskiy-ICLR-2021}                              
& & \cmark & 0.559& 0.137 & 0.832 & 0.754 \\
CLIP Image \cite{Radford-ICML-2021}                       
& & \cmark & 0.549 & 0.523 & 0.822 & 0.746 \\
\hline
DistilBERT+ViT (Teacher 1)           
& \cmark & \cmark & 0.920 & 0.919 & 0.867$^\ddagger$ & 0.843 \\
DistilBERT+CLIP (Teacher 2)           
& \cmark & \cmark & 0.931$^\ddagger$ &  0.926$^\ddagger$ & 0.868$^\ddagger$ & 0.846 \\
\hline
DistilBERT (Student 1) 
& \cmark & & 0.930$^\ddagger$ & { 0.928}$^\ddagger$ & {0.869$^\ddagger$ }& {0.843} \\
DistilBERT (Student 2) 
& \cmark & & {\bf 0.931}$^\ddagger$ & {\bf 0.929}$^\ddagger$ & {\bf 0.871}$^\ddagger$ & {\bf 0.848}$^\ddagger$ \\
\hline
\end{tabular}
\vspace{-0.3cm}
\end{table}

\noindent
\textbf{Baselines and backbones.} As baseline, we choose the DistilBERT model \cite{Sanh-EMC-2019}, a variant of BERT \cite{Devlin-NAACL-2019} that exhibits good performance with a reasonable number of learnable parameters. For a fair comparison with the baseline, we employ the DistilBERT architecture for our students as well. Moreover, the text encoder inside the multimodal teachers is also based on DistilBERT. To encode the generated images, we alternatively employ the pre-trained image encoder of the CLIP architecture \cite{Radford-ICML-2021}, or the pre-trained ViT \cite{Dosovitskiy-ICLR-2021} model. We thus obtain a teacher based on DistilBERT+ViT (Teacher 1), and a teacher based on DistilBERT+CLIP (Teacher 2). We distill the knowledge from Teacher 1 into a student based on DistilBERT (Student 1), and the knowledge from Teacher 2 into a different student (Student 2), which is also based on DistilBERT. We underline that the two students have the same architecture, but they differ in terms of the source providing the privileged information.

\noindent
\textbf{Hyperparameters.}
We train the models with the AdamW \cite{Loshchilov-ICLR-2019} optimizer using a learning rate of $5 \cdot 10^{-5}$ with linear decay, which converges to good optima across all our experiments. The baseline DistilBERT, the teachers and the students are each trained for 100 epochs on an Nvidia GeForce GTX 1080Ti GPU with 11 GB of VRAM. In all the experiments, we use a mini-batch size of 14 samples. Following previous works on knowledge distillation \cite{Ba-NIPS-2014,Lopez-ICLR-2016}, we soften the output of the teacher using the temperature $\tau$. We validate this hyperparameter in the range $1$-$10$, achieving optimal results with $\tau=8$. The hyperparameters $\alpha$ and $\beta$ from Eq.~\eqref{eq_loss_KD} are validated in the range from $0.1$ to $5$. The optimal values are $\alpha=3$ and $\beta=1$. 
        
\noindent
\textbf{Data preprocessing.}
Before generating images with Stable Diffusion v2 \cite{Rombach-CVPR-2022}, we perform some preprocessing steps to clean up the text samples. For the IMDB data set, we remove the HTML tags that are sometimes present in movie reviews. For the 20 Newsgroups data set, we discard email addresses and subjects, using the remaining content as text prompt. For the English News and English WikiNews data sets, we provide the target word or multi-word phrase in each sentence as input for the text-conditional diffusion model. This is because the task is to identify the complexity of the target words, not of the whole sentences.

To process the examples from the English News and English WikiNews corpora with DistilBERT, we modify each sentence by marking the target words or multi-word phrases with the \emph{[SEP]} token. No further preprocessing is required for the other data sets.

\vspace{-0.2cm}
\subsection{Results}
\vspace{-0.2cm}

We present the results obtained on the IMDB, 20 Newsgroups, English News and English WikiNews data sets in Table~\ref{tab1}.

\noindent
\textbf{IMDB.} 
The baseline DistilBERT model \cite{Sanh-EMC-2019}, which is trained using only text data, reaches an accuracy of $91.9\%$, while the image encoders barely surpass the random chance baseline. The best multimodal teacher employing the CLIP image encoder reaches an accuracy of $93.1\%$. Our first student outperforms its teacher by $1\%$, while our second student is on par with its teacher. Notably, both students surpass the baseline model by more than $1.1\%$.

\noindent
\textbf{20 Newsgroups.}
The baseline DistilBERT \cite{Sanh-EMC-2019} obtains a performance of $91.8\%$, while the individual image encoders lag far behind. Since ViT is much worse than CLIP, the corresponding teacher (DistilBERT+ViT) barely surpasses the baseline model, while DistilBERT+CLIP (Teacher 2) reaches an accuracy of $92.6\%$. Meanwhile, our students based on privileged information surpass their teachers, showing considerable performance gains over the baseline DistilBERT.

\noindent
\textbf{English News.} 
On the English News corpus, the baseline DistilBERT obtains an accuracy of $86.1\%$. The ViT and CLIP image encoders obtain competitive results, being less than $4\%$ behind DistilBERT \cite{Sanh-EMC-2019}. Both multimodal teachers outperform the baseline DistilBERT. Moreover, our student models surpass their teachers. The best student outperforms the baseline DistilBERT by $1\%$, reaching an accuracy of $87.1\%$ in complex word identification.

\noindent
\textbf{English WikiNews.} The results on the English WikiNews corpus are consistent with those on the English News corpus. Indeed, the independent image encoders obtain fairly good results, given that they only take generated images as input. The multimodal teachers outperform the baseline DistilBERT, while the students yield even better results.

\begin{figure}[!th]
    \centering 
    \includegraphics[width=1.0\linewidth]{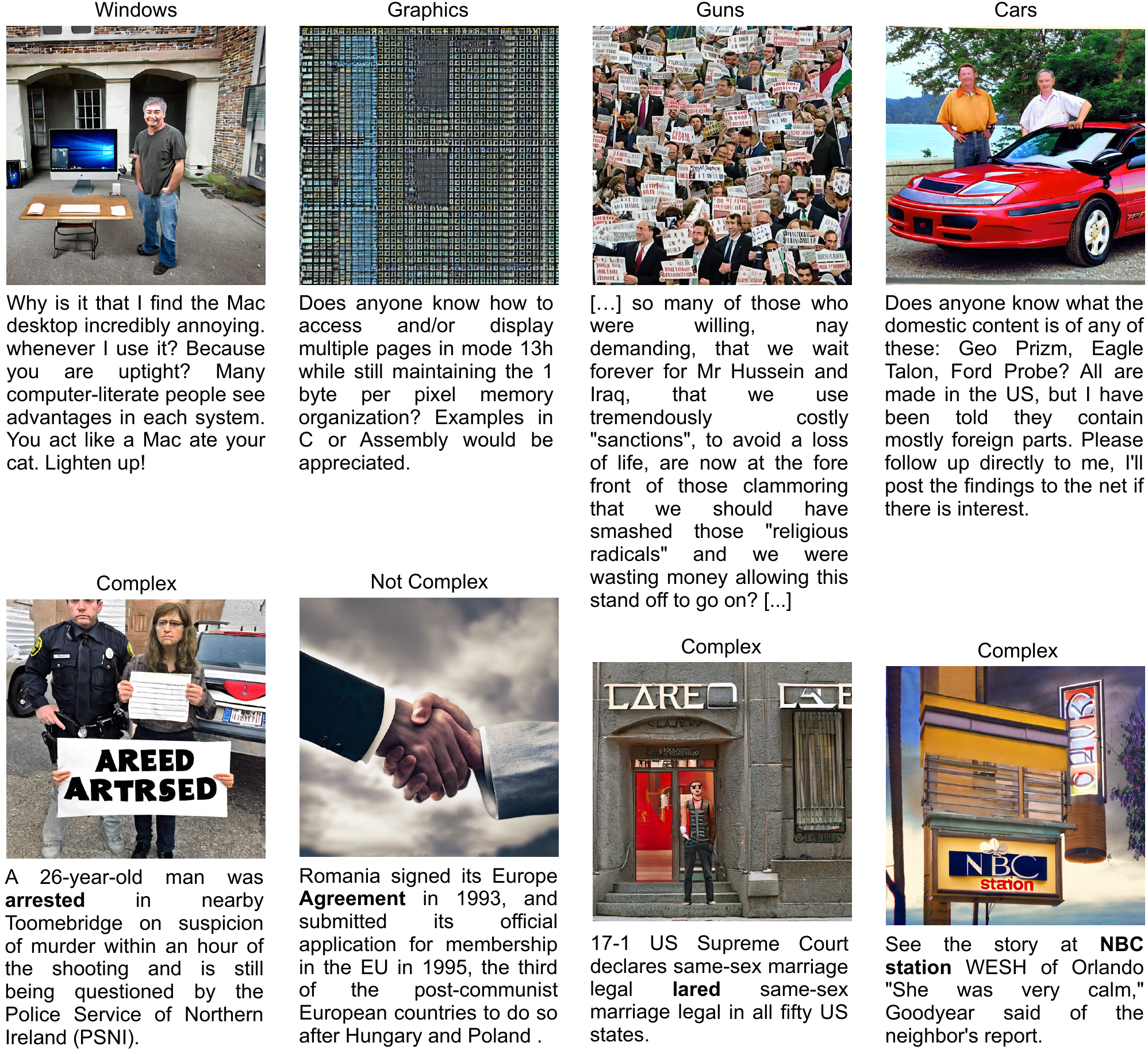}
    \vspace{-0.5cm}
        \caption{Text samples and generated images that are correctly classified by the multimodal teacher based on  DistilBERT+CLIP. The target label is displayed on top of each sample. The examples on top belong to the 20 Newsgroups~\cite{Lang-ICML-1995} data set, while the examples below are taken from English News \cite{Yimam-RANLP-2017} and English WikiNews \cite{Yimam-RANLP-2017}.\vspace{-0.4cm}}
        \label{fig:examples}
\end{figure}

\noindent
\textbf{Overall.}
We notice that the text modality leads to better results than the image modality, regardless of the data set. This is a natural consequence of the fact that the images are generated by a diffusion model, which can produce images that do not reflect the label. Another generic observation is that the multimodal teacher based on the CLIP image encoder (Teacher 2) is generally better than the other teacher. This leads to a better DistilBERT student (Student 2). Furthermore, we observe that the students generally surpass their teachers. We explain this observation through the fact that the multimodal teachers assign equal importance to the text and image modalities, although the image modality is naturally inferior. In contrast, the students focus on the original text modality, obtaining information about the image modality only through knowledge distillation.

Since both students surpass the baseline DistilBERT in each and every case, we conclude that our LUGPI framework is beneficial in various text classification tasks, such as polarity classification, text categorization by topic, and complex word identification.

\noindent
\textbf{Qualitative results.}
In Figure~\ref{fig:examples}, we illustrate some examples which are incorrectly classified by the baseline DistilBERT, but are correctly classified by our second teacher model (DistilBERT+CLIP). Remarkably, we observe that the images generated by Stable Diffusion v2 contain important clues. For instance, a car is generated when the prompt is about cars, even though the word ``car'' is never mentioned inside the prompt. For the complex word identification task, we observe that the images generated for  simple (non-complex) words tend to be less abstract, while those generated for complex words tend to be more abstract. In summary, the illustrated examples show that the generated images can complement the corresponding text samples. Although our students do not see these images at test time, our quantitative results presented in Table~\ref{tab1} show that the students clearly benefit from the privileged information transferred from the multimodal teachers.

\begin{table}[!t]
\caption{Accuracy rates on IMDB Large Movie Review \cite{Maas-ACL-2011}, 20 Newsgroups \cite{Lang-ICML-1995}, English News \cite{Yimam-RANLP-2017} and English WikiNews \cite{Yimam-RANLP-2017} data sets, while ablating the knowledge distillation components of our loss defined in Eq.~\eqref{eq_loss_KD}. The best accuracy on each corpus is highlighted in bold.}
\label{tab2}
\centering
\renewcommand{\arraystretch}{1.18}
\begin{tabular}{|l|c|c|c|c|c|c|}
\hline
\multirow{2}{*}{Model} & \multicolumn{2}{c|}{Loss Terms} & {IMDB} & {20 News} & English & English\\
\cline{2-3}
     & $\;\!\mathcal{L}^T_{\scriptsize{\mbox{CE}}}\;\!$ & $\mathcal{L}^T_{l_2}$ & Reviews & groups & News & WikiNews\\
\hline
\hline
DistilBERT (Student 1) & & & 0.919 & 0.918 & 0.861 &  0.842 \\
DistilBERT (Student 2) & & & 0.919 & 0.918 & 0.861 &  0.842 \\
\hline
DistilBERT (Student 1) & \cmark & & 0.913 & 0.922 & 0.765 & 0.842 \\
DistilBERT (Student 2) & \cmark & & 0.923 & 0.926 & 0.870 & 0.844 \\
\hline
DistilBERT (Student 1) &  & \cmark & 0.911 & 0.926 & 0.869 & 0.840 \\
DistilBERT (Student 2) &  & \cmark & 0.919 & 0.925 & 0.865 & 0.843 \\
\hline
DistilBERT (Student 1) & \cmark & \cmark & 0.930 & { 0.928} & {0.869}& {0.843} \\
DistilBERT (Student 2) & \cmark & \cmark & {\bf 0.931}& {\bf 0.929} & {\bf 0.871}& {\bf 0.848} \\
\hline
\end{tabular}
\vspace{-0.3cm}
\end{table}

\noindent
\textbf{Ablation study.} Our LUGPI framework performs the distillation at two network levels, via two distinct loss terms. To demonstrate the utility of both terms, we perform an ablation study of the knowledge distillation loss terms $\mathcal{L}^T_{\scriptsize{\mbox{CE}}}$ and $\mathcal{L}^T_{l_2}$ from Eq.~\eqref{eq_loss_KD}. We present the corresponding results in Table \ref{tab2}. Distilling knowledge at the output level via $\mathcal{L}^T_{\scriptsize{\mbox{CE}}}$ is not beneficial for the first student. In contrast, distilling knowledge at the embedding level via $\mathcal{L}^T_{l_2}$ helps both students on three data sets (except IMDB). In summary, the ablation study shows that both distillation losses are required to obtain consistent improvements.

\noindent
\textbf{Training and inference time.}
The inference time of our final model is identical to that of the vanilla DistilBERT. However, the training time of our pipeline is between $2.3\times$ and $2.8\times$ higher (depending on the dataset and the vision model) than that of the student. This includes the time for generating the images with the pre-trained Stable Diffusion model. Note that Stable Diffusion is kept frozen in our pipeline.

\vspace{-0.2cm}
\section{Conclusion}
\vspace{-0.2cm}

In this work, we proposed the Learning Using Generated Privileged Information framework, which employs a diffusion model to generate privileged images, which were further used to train a multimodal teacher taking both text and image data as input. A unimodal student was subsequently trained by distilling privileged information from the multimodal teacher. We performed experiments on four text classification data sets, namely IMDB Movie Reviews, 20 Newsgroups, English News and English WikiNews. We alternatively employed two different image encoders to extract image features, demonstrating accuracy gains in both cases. All our distilled students outperformed the baseline model and even the multimodal teachers, without any extra cost during inference. In future work, we aim to extend our framework to more NLP tasks.

\bibliographystyle{splncs04}
\bibliography{references}

\end{document}